\documentclass[10pt,twocolumn,letterpaper]{article}

\usepackage{iccv}
\usepackage{times}
\usepackage{epsfig}
\usepackage{graphicx}
\usepackage{amsmath}
\usepackage{amssymb}

\usepackage{verbatim}
\usepackage{multirow}
\usepackage{booktabs}
\usepackage{xcolor}
\usepackage{tablefootnote}

\usepackage[pagebackref=true,breaklinks=true,letterpaper=true,colorlinks,bookmarks=false]{hyperref}

\newcommand{\head}[1]{\noindent\textbf{#1}}
\newcommand{\cupid}{task-adaptive video-language pre-training } 

\newcommand{\cupidabbr}{\textsc{Cupid} }

\iccvfinalcopy 

\def\iccvPaperID{7069} 

\ificcvfinal\fi

\begin{document}

\title{\textsc{Cupid}: Adaptive Curation of Pre-training Data for Video-and-Language Representation Learning} 

\author{Luowei Zhou, Jingjing Liu, Yu Cheng, Zhe Gan, and Lei Zhang \\
Microsoft Cloud and AI \\
{\tt\small \{luozhou, jingjl, yu.cheng, zhe.gan, leizhang\}@microsoft.com}
}

\maketitle

\begin{abstract}
This work concerns video-language pre-training and representation learning. In this now ubiquitous training scheme, a model first performs 
pre-training on paired videos and text (e.g., video clips and accompanied subtitles)
from a large uncurated
source corpus, before transferring to specific downstream tasks. This two-stage training process inevitably raises questions about the generalization ability of the pre-trained model, which is particularly pronounced when a salient domain gap exists between source and target data (e.g., instructional cooking videos vs. movies). In this paper, we first bring to light the sensitivity of pre-training objectives (contrastive vs. reconstructive) to
domain discrepancy. Then, we propose a simple yet effective framework, \textsc{Cupid}, to bridge this domain gap by filtering and adapting source data to the target data, followed by domain-focused pre-training. Comprehensive experiments demonstrate that pre-training on a considerably small subset of domain-focused data
can effectively close the source-target domain gap and achieve significant performance gain, compared to random sampling or even exploiting the full pre-training dataset.
\textsc{Cupid} yields new state-of-the-art performance across multiple video-language and video tasks, including text-to-video retrieval~\cite{ZhXuCoAAAI18, lei2020tvr}, video question answering~\cite{lei2018tvqa}, and video captioning~\cite{ZhXuCoAAAI18}, with consistent performance lift over different pre-training methods.
\end{abstract}

\section{Introduction}



Recent advances in contrastive learning for computer vision~\cite{oord2018representation,he2020momentum,chen2020simple} and language model pre-training in natural language processing~\cite{devlin2018bert,liu2019roberta,brown2020language} have brought forth a new wave of innovations on self-supervised representation learning. Vision and language, used to be widely disparate fields, have started to exhibit considerable similarities and synergy, considering the adoption of convolution networks for language modeling~\cite{kim-2014-convolutional} and Transformer for image classification~\cite{dosovitskiy2020image}. This inspired a line of research~\cite{sun2019videobert,lu2019vilbert,chen2020uniter,zhou2020unified,miech2019howto100m,zhu2020actbert,miech2020end,radford2021learning,jia2021scaling} in the intersection of vision and language, encapsulated in large-scale multimodal pre-training with paired textual and visual content, which captures the essence of self-supervision from both sides. Video, in particular, contains both complex visual information and rich semantic information from speech that are easy to harvest from the wild at a massive scale, providing a fertile field for multimodal studies. 

\begin{figure}[t!]
\centering
   \includegraphics[width=1.\linewidth]{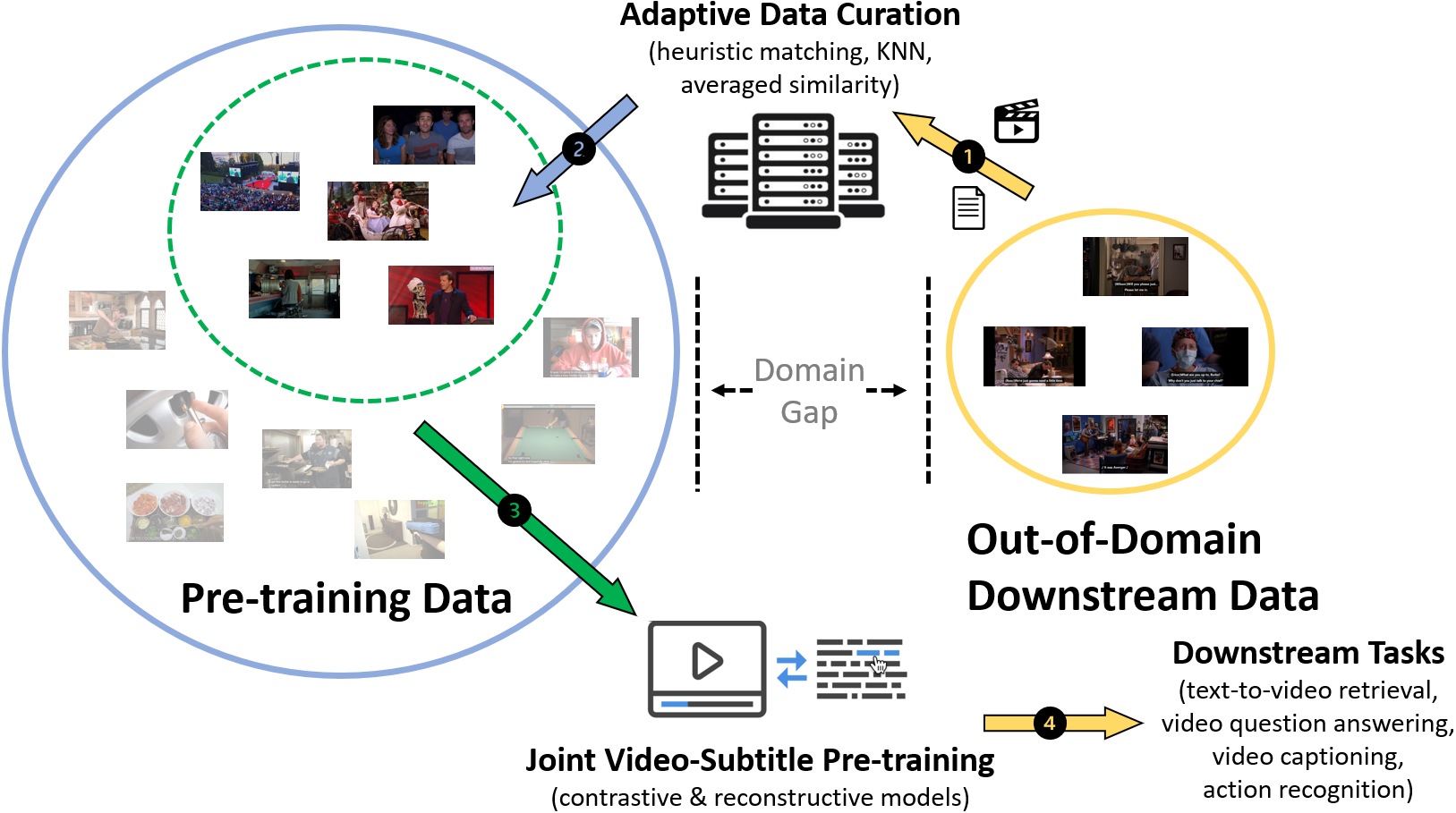}
   \caption{A diagram illustration of our proposed adaptive data curation method for video-language pre-training. Assume we have a large-scale \emph{source} dataset for pre-training and a \emph{target} dataset from downstream tasks. We propose a pre-training strategy to close the potential domain gap between source and target. We first filter the source dataset for videos visually similar to the target data. Three strategies are introduced for this purpose, namely, heuristic matching, K-Nearest Neighbours, and averaged similarity score, where video content or metadata are used for determining the similarity between videos. The resulted small subset of the source data is then used to pre-train a video-language model that could later be finetuned for various downstream tasks.}
\label{fig:tvqa_c}
\end{figure}

Existing pre-training methods often assume ready access to a large corpus with paired videos and captions (\eg, video clips and accompanied subtitles).
The training scheme comprises of two stages: a model first performs either contrastive or reconstructive pre-training on paired videos and text, and then the pre-trained model is transferred to specific downstream tasks. This two-stage training process inevitably raises questions about the generalization ability of the pre-trained model, which is particularly pronounced when a salient domain gap exists between source and target data (\eg, instructional cooking videos \vs movies).
In this paper, we aim to answer the following questions. Does the prevailing pretrain-then-finetune paradigm imbue vision-and-language models with strong generalization ability? 
If there exists a domain gap between the pre-training and downstream dataset, how can we close the gap efficiently?
We start our quest by first rolling out an overview of existing video datasets (Figure~\ref{fig:video_dataset}). Under this framing, 
we conduct a probing analysis and propose a measurement for pre-training transferability that is faithful to the pre-training objective.
From this analysis, we reveal a major performance degradation 
when pre-trained models are transferred to out-of-domain data.


To this end, we propose \textsc{Cupid} (\textbf{CU}ration of \textbf{P}re-tra\textbf{I}ning \textbf{D}ata), a simple yet effective framework for video-language pre-training. Our goal is to filter the pre-training source dataset to select a small portion of videos similar to downstream target datasets so as to eliminate domain discrepancy, if any.
Training on such focused dataset is faster than using the full dataset for pre-training, as well as yielding better pre-training transferability thanks to better domain match. As during \emph{full} pre-training, even though the same data have been seen by the model,  they can be diluted by the massive amount of ``distractors'' from other distant domains. Therefore, the aim of our proposed strategies is to enable the model to automatically discover the ``gold nuggets'' from the enormous noisy source dataset, followed by a laser-focused pre-training on domain data. 

To fulfill the data curation process, we propose two strategies that measure the similarity between source and target videos, either by measuring the distance between two videos in the embedding space, or the relevance of video metadata such as category and title.
We comprehensively evaluate the proposed adaptive pre-training method over state-of-the-art (SotA) video-language pre-trained models, across a variety of tasks including text-to-video retrieval~\cite{lei2020tvr,ZhXuCoAAAI18}, video captioning~\cite{ZhXuCoAAAI18}, and video question answering~\cite{lei2018tvqa}.
\textsc{Cupid} achieves new SotA performance on these tasks 
and gains surprising superior performance even over full pre-training (with 80\% fewer data) on some of these tasks.

Our contributions are summarized as follows.
    ($i$) We propose \textsc{Cupid}, a \cupid framework that effectively closes the domain gap between source (for pre-training) and target (for finetuning) data.
    ($ii$) We provide a probing study to examine domain sensitivity over contrastive and reconstructive pre-training objectives.
    ($iii$) We provide a comprehensive evaluation of \textsc{Cupid} on a wide range of video-language and video tasks, with consistent improvement and SotA performance.

\section{Related Work}
Self-supervised video representation learning has been widely applied, from predicting future sequence~\cite{ranzato2014video,srivastava2015unsupervised}, to sorting or classifying correct frame order~\cite{fernando2017self,lee2017unsupervised}, and most recently to contrastive learning~\cite{qian2020spatiotemporal,han2020self}. Universal language model pre-training has emerged in recent years with notable success such as BERT~\cite{devlin2018bert}, RoBERTa~\cite{liu2019roberta} and GPT-3~\cite{brown2020language}. At their intersection, video-language representation learning takes inspiration from both worlds and has become a nascent research area with growing attention from both vision and NLP community.

The learning objectives of existing methods largely fall into two  categories: contrastive and reconstructive.
The first efforts on contrastive video-language pre-training include \cite{miech2019howto100m} and \cite{sun2019learning}, where ranking loss or noise contrastive estimation (NCE) is used to distinguish paired video-subtitle data with distractors while learning joint video-language representation. One limitation of this approach is that the paired video clip and subtitle might not correspond to each other semantically, due to potential misalignment between the visual scene and semantics in the subtitle. A follow-up work named MIL-NCE~\cite{miech2020end} addresses this issue by multiple instance learning (MIL), where a video clip is associated with multiple neighbouring subtitles instead of one instance to facilitate better vision-language grounding.

Reconstructive methods heavily rely on masked language modeling (MLM) which is originated from language pre-training BERT~\cite{devlin2018bert} and its visual counterpart masked region modeling (MRM). VideoBERT~\cite{sun2019videobert} converts video frames to tokens through clustering, so a joint masked video-text modeling can fit into a MLM framework. ActBERT~\cite{zhu2020actbert} introduces masked action classification and masked object classification, which leverages fine-grained language semantics for self-supervised learning. Li \etal propose HERO~\cite{li2020hero}, which relies on reconstructive proxy tasks such as frame order prediction and masked frame modeling (it also resorts to contrastive objective such as video subtitle matching). Another work UniVLM~\cite{luo2020univilm} introduces a language decoder to reconstruct masked input in an auto-regressive fashion. Besides language, other modalities such as audio and speech~\cite{alayrac2020self,rouditchenko2020avlnet,alwassel2019self} are also studied. 

In this paper, we propose a generic framework that can be readily applied to any of the methods aforementioned. We select the best-performing model from each category as the surrogate in our experiments, but similar analysis can be applied to other models. To the best of our knowledge, we are the first to systematically study the domain gap issue in video-language pretrain-then-finetune frameworks. Even though \cite{miech2019howto100m} demonstrates that the representation learned from pre-training instructional videos can generalize across domains to movies after task-specific finetuning, we show that this is not enough. Domain gap is still persistent after pre-training and if we close the gap, the finetuned models get further boosted. A related work~\cite{singh2020we} observes that pre-training on generated texts in a domain similar to the downstream task (\eg, questions for VQA~\cite{goyal2017making}) is better than pre-training on natural data but from a different domain (\eg, captions from Conceptual Captions~\cite{sharma2018conceptual}). Our method shares some merit with a recent trend in the NLP field on domain-specific pre-training~\cite{gururangan2020don,shin-etal-2020-biomegatron}, but we focus on a multimodal setting with an in-depth analysis on how video models react to domain gap. 
Due to space limit, we review downstream tasks
in the Appendix.

\section{Understanding the Domain Gap}
In this section, we start with an overview of video understanding landscape. 
This cartography exercise  and its consequent observations are followed by a probing analysis on domain gap and its impact on pre-training transferability.

\subsection{Domain Cartography}
\begin{figure}[!t]
\begin{center}
   \includegraphics[width=1.\linewidth]{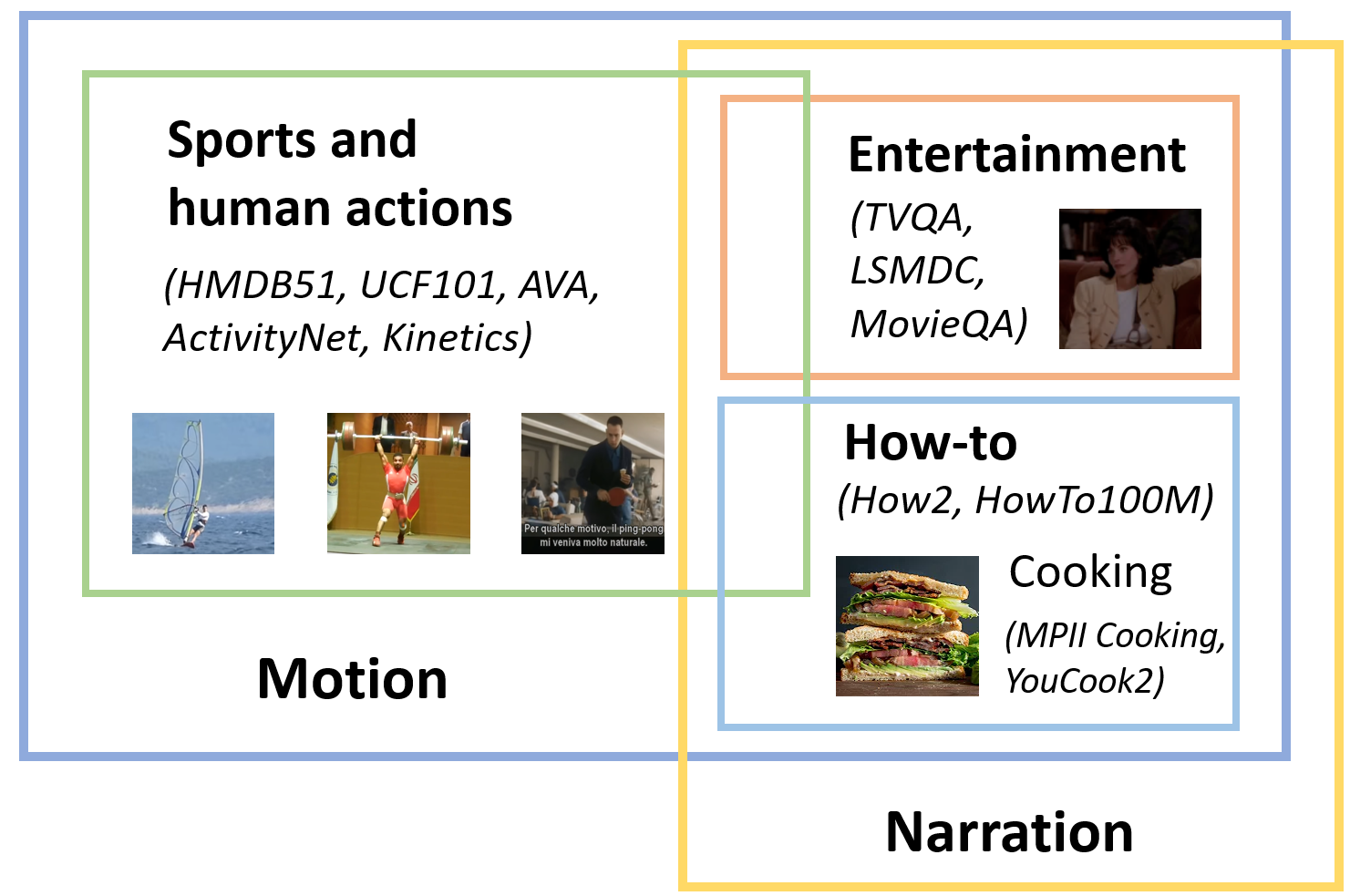}
\end{center}
   \caption{The landscape of video understanding datasets.} 
\label{fig:video_dataset}
\end{figure}

Existing video datasets largely fall into two camps of content: \emph{Motion} and \emph{Narration}, as illustrated in Figure~\ref{fig:video_dataset}. Motion videos are visually dynamic and often exude complex motion flows and transitions. Narration videos, on the other hand, are more static and often contain rich monologue narrations. Note that these two types are not mutually exclusive; for instance, instructional videos with narrations sometimes involve human activities, even though monotone (Figure~\ref{fig:video_dataset}). Under this umbrella, different \textit{domains} of videos have been widely studied in the community, such as \emph{sports and human actions} videos (\eg, HMDB51~\cite{kuehne2011hmdb}, UCF101~\cite{soomro2012ucf101}, ActivityNet~\cite{caba2015activitynet}, AVA~\cite{gu2018ava}, and Kinetics~\cite{carreira2017quo}), \emph{entertainment} videos (\eg, MovieQA~\cite{tapaswi2016movieqa}, LSMDC~\cite{lsmdc}, TVQA~\cite{lei2018tvqa}, and TVR~\cite{lei2020tvr}), and \emph{how-to} videos (\eg, MPII Cooking~\cite{rohrbach2012database}, Narrated Videos~\cite{alayrac2016unsupervised}, YouCook2~\cite{ZhXuCoAAAI18}, Epic-Kitchens~\cite{damen2018scaling}, How2~\cite{sanabria2018how2}, HowTo100M~\cite{miech2019howto100m}, COIN~\cite{tang2019coin}, and CrossTask~\cite{zhukov2019cross}).

This ontology can be further combed down to a finer-grained level, \eg, \emph{cooking} videos as one sub-domain of \emph{how-to} videos (Figure~\ref{fig:video_dataset}). Note that the taxonomy here is by no means comprehensive, with notable exceptions such as driving time-lapses (\eg, KITTI~\cite{Geiger2012CVPR} and Cityscapes~\cite{Cordts2016Cityscapes}) and surgical recordings (\eg, JIGSAWS~\cite{gao2014jhu} and m2cai16-tool~\cite{twinanda2016endonet}). 
For simplicity, our discussion focuses on the domains outlined earlier as mainstream benchmarks, and leaves other orphan categories to future study.

Domain gap has been a consistently challenging issue in video understanding studies. Due to drastic disparity between different video domains (\eg, sports vs. cooking videos), a well-trained model learned from one domain often scales poorly to another video domain if applied directly. With the discussed hierarchical domain cartography in mind, we search for a systematic and simple framework that can capture different cases of domain gap, so that it can consequently guide us to better model design for domain generalization. To this end,  given a pivotal domain, we define three general types of domain relations: 
\begin{itemize}
\item \head{In-domain:} Video types are the same as or a subset of the pivotal domain;
       \item \head{Near-domain:}Video types overlap with the pivotal domain to some extent;
      \item \head{Out-of-domain:} Video types barely overlap with the pivotal domain.
\end{itemize}

Given that HowTo100M~\cite{miech2019howto100m} is by far the largest video dataset, including 120M paired video clips and ASR subtitles on instructional videos, we use this large-scale dataset as our pivotal domain, or \textbf{source dataset}, in our investigation. We select one representative dataset from each general category (\emph{sports and human actions}, \emph{entertainment}, \emph{how-to}) that are widely studied in the literature, and use them as the \textbf{target datasets} in our analysis.  



\head{In-domain Datasets.} \textbf{YouCook2}~\cite{ZhXuCoAAAI18} is considered as in-domain, which consists of 2000 untrimmed cooking videos sourced from YouTube. 
Recipe steps in each video are localized temporally with start and end timestamps and described by a caption. We treat each video clip as a standalone video and conduct standard evaluations on two tasks: video captioning and text-to-video retrieval (or retrieval for simplicity). Since the test set annotation is unavailable, results are reported on the validation set.

\head{Near-domain Datasets.} 
\textbf{HMDB51}~\cite{kuehne2011hmdb} is a dataset for human action recognition. It contains 7K clips distributed across 51 human action classes. The goal is to classify a given video into one of the pre-defined action classes. 

\head{Out-of-domain Datasets.} \textbf{TVQA}~\cite{lei2018tvqa} and \textbf{TVR}~\cite{lei2020tvr} 
datasets are based on the TV dataset~\cite{lei2018tvqa} mentioned earlier and have identical train, val, and test splits. TV dataset is considered as out-of-domain which has no overlap with Howto100M videos. In TVQA, the goal is to answer a multiple-choice question about a given video. A temporal grounding (locating start and end timestamps) of the question is provided as a part of the annotation. TVR dataset concerns the problem of video corpus moment retrieval where given a query caption, a model needs to retrieve the most relevant video from the corpus and temporally ground the query in the retrieved video. 

Standard recall metrics are used for text-to-video retrieval assessment. The overall accuracy is reported on QA and action recognition tasks. Standard language matching metrics including BLEU, METEOR, CIDEr, and sometimes ROUGE-L are reported for captioning.\footnote{Note that for captioning on YouCook2, current evaluation diverges into two modes: micro-averaging and macro-averaging. \href{https://github.com/tylin/coco-caption}{Micro-averaging} indicates that metric scores are averaged across all segments. \href{https://github.com/ranjaykrishna/densevid_eval}{Macro-averaging} indicates that metric scores are first averaged within each (long) video and then averaged across the entire split. For completeness and fair comparison, we include both modes.} 
Note that datasets that have the same distribution as the source domain (\eg, How2R/How2QA~\cite{li2020hero}) or from an open domain (\eg, ~\cite{xu2016msr,li2016tgif}) are not considered in this work as the domain gap is not applicable and a generic pre-training is sufficient.

\subsection{Pre-training Transferability}\label{sec:pretrain_transfer}
The de-facto approach to evaluating pre-trained model is through finetuning on downstream tasks and evaluating on task-specific metrics. However, two main caveats advise caution. 
First, downstream finetuning introduces intractable factors to the training process (\eg, the auxiliary layer(s) for downstream tasks~\cite{singh2020we,miech2019howto100m}, task-specific hyper-parameter search~\cite{miech2019howto100m}), which impedes the assessment of general pre-training quality over a common ground. Second, metrics on different tasks might not be comparable, and finetuning on individual tasks takes significant amount of time. To gain insights on how source-target domain gap impacts the transferability of a pre-trained model, we conduct a probing analysis by evaluating pre-training objectives directly, and measuring their performance on downstream data (only data, hence task-agnostic). 

In popular pre-trained frameworks, both reconstructive (such as BERT) and contrastive learning objectives (such as NCE) are adopted. 
Here, we choose MLM (masked language modeling) accuracy as the transferability measurement for reconstructive objectives, and 
zero-shot text-to-video retrieval accuracy for contrastive objectives. We adopt HERO model architecture for MLM training, and MIL-NCE for contrastive training.




As aforementioned, we use HowTo100M as the source dataset in our analysis. For MLM, we choose TV data~\cite{lei2018tvqa} as our out-of-domain target dataset. When applying a model pre-trained with Howto100M training set (\emph{Model I}) to HowTo100M val set, the accuracy of recovering masked word is 54.51\%. When the model is pre-trained on TV training set (\emph{Model II}) and evaluated on TV val set, accuracy is 69.93\%. Both models show relatively good generalization, as training/val are from the same video type. But when  Model I is directly applied to TV val set, accuracy drops to 31.17\%, indicating significant domain discrepancy.



Then, we evaluate an MIL-NCE model pre-trained on HowTo100M over two datasets: YouCook2 (\emph{cooking}) and TV. 
On the in-domain dataset YouCook2\footnote{Subtitles for YouCook2 obtained from Azure Speech to Text Service.}, we observe 5.8\%/13.8\%/19.2\% on Recall@1/@5/@10. On the out-of-domain dataset TV, the accuracy is drastically low (0.49\%/1.93\%/3.50\%), indicating again a pronounced domain gap. We later show that our proposed method in Section~\ref{sec:tqvlp} closes this gap significantly, achieving superior performance over current state of the art.



\begin{table*}[!pt]
\centering
\small
\begin{tabular}{llccccccccc}
\toprule
\multirow{2}{*}{Method} & \multirow{2}{*}{Curated PT Data} & \multicolumn{4}{c}{YouCook2 (Captioning)} & \multicolumn{4}{c}{YouCook2 (Retrieval)} & HMDB51 \\ \cline{3-11} 
 & & B@3 & B@4 & M & C & R@1 & R@5 & R@10 & mR ($\downarrow$) & Accu. \\
 \midrule
Baseline & None & 11.57 & 7.21 & 13.92 & 87.46 & 15.80 & 40.34 & 54.10 & 8 & 63.40 \\ 
 & 15K HT100M (Random) & 12.67 & 7.92 & 15.18 & 98.67 & 15.89 & 41.23 & 54.14  & 8 & 64.23 \\ 
 \midrule
\textsc{Cupid} (Ours) & 15K HT100M (Heuristic) & \textbf{13.93} & \textbf{8.83} & \textbf{16.10} & \textbf{105.94} & \textbf{17.67} & \textbf{43.21} & \textbf{57.06} & \textbf{7} & \textbf{65.45} \\ 
\bottomrule
\end{tabular}
\vspace{2mm}
\caption{Results on in-domain/near-domain tasks with HAP curation strategy. For YouCook2 captioning~\cite{ZhXuCoAAAI18}, we adopt the ``Micro-Averaging'' evaluation setting (B: BLEU, M: METEOR, C: CIDEr). mR in retrieval indicates the median recall. Linear evaluation is conducted for action recognition on HMDB51~\cite{kuehne2011hmdb}, with accuracy averaged across all three splits.}\label{tab:hap_results}
\end{table*}


\section{Adaptive Pre-training via Data Curation}\label{sec:tqvlp}

With these various video domains in mind, our goal is to introduce an adaptive data curation strategy to filter source data into a selective, much smaller, subset of data that is most relevant to any given target domain.
We introduce two simple yet highly effective methods for adaptive data curation, with the goal of locating a small portion of videos in the source dataset that are visually and semantically similar to target domain: ($i$) heuristic-based; and ($ii$) similarity-based. 

\subsection{Heuristic Adaptive Pre-training (HAP)}\label{sec:hap}
First, we start with a straightforward heuristic method, to tackle in-domain/near-domain data curation. The meta information of video, such as video category and title, serves as informative indicator of semantic content in the video. For instance, in HowTo100M, 1.3\% of data belongs to the pre-defined category of ``Sports and Fitness'', and 40.9\% belongs to ``Food and Entertaining''. To leverage this given guidance, we design a host of heuristic rules to rely on video metadata for effective subset curation. For example, the video category of ``Sports and Fitness'' in source data is selected for downstream tasks in domain genre \emph{sports and human actions}, and the video category of ``Food and Entertaining'' is selected for downstream domain genre \emph{cooking}. We further shrink the candidate pool by constraining the source video title to contain at least one word that exists in the metadata of downstream video
and have user-uploaded subtitles (roughly 10\% of YouTube videos are accompanied with human-generated subtitles; the rest 90\% are ASR-generated).
In this way, we collect around 15K videos for \emph{sports and human actions} and 15K for \emph{cooking}
(\ie, 1.2-1.3\% of the original HowTo100M dataset). 

\paragraph{Results on In-domain/Near-domain Tasks.}
Results on adaptive pre-training with heuristic data curation are presented in Table~\ref{tab:hap_results}, including in-domain data YouCook2 and near-domain data HMDB51. Two tasks on YouCook2 are considered: video captioning and text-to-video retrieval. The goal of video captioning is to describe a given video clip with a natural language sentence. Text-to-video retrieval aims to retrieve a video from a set of videos that is the most relevant to the query text. For HMDB51, we perform standard action recognition task with \emph{linear evaluation} to assess the learned visual representation. The model architecture is based on VLP~\cite{zhou2020unified} and CLIP~\cite{radford2021learning} (details in Section~\ref{sec:sota}). From the table, we see that our proposed method HAP outperforms the random sampling baseline on all metrics considered. Particularly on YouCook2 in-domain tasks, HAP demonstrates a clear advantage over the baseline, with 5-11.5\% relative improvements across-the-board. It is worth noting that on the retrieval task, we see little-to-no improvement from the light 15K pre-training data from random sampling, but a major boost with our heuristic sampling strategy. On the near-domain task, we observe that pre-training with randomly sampled data leads to marginal improvement, possibly due to the domain shift. The effectiveness of HAP remains sound, with a 3\% boost over the model without pre-training.

\subsection{Similarity-based Adaptive Pre-training (SAP)}\label{sec:sap}
HAP is designed for in-domain and near-domain target data. For out-of-domain data such as TV and movies, this approach is not applicable as few videos from HowTo100M are labeled under this category.
To accommodate more challenging domain discrepancy in out-of-domain tasks, we introduce the following similarity-guided curation.

Given a specific downstream task, our goal is to efficiently select a small subset from the source dataset that is visually similar to the target domain, therefore most effective in closing the domain gap between pre-training and the downstream task at hand. Formally, we denote the \textit{source dataset} for pre-training and \textit{target dataset} for downstream task as $V_\mathcal{S}=\{v_\mathcal{S}^1, v_\mathcal{S}^2, \ldots, v_\mathcal{S}^N\}$ and $V_\mathcal{T}=\{v_\mathcal{T}^1, v_\mathcal{T}^2, \ldots, v_\mathcal{T}^P\}$, respectively, where $v_\mathcal{S}^i$ and $v_\mathcal{T}^j$ are the video clips with index $i$ and $j$, and $N$ and $P$ are the total number of videos. Note that $N\gg P$. 
Our objective is to filter $V_\mathcal{S}$ for videos similar to $V_\mathcal{T}$, to compensate domain discrepancy between source and target. An intuitive first step towards effective subset selection is to rely on visual similarity between videos as a domain relevance measurement to guide our task-adaptive data curation.

Given a video from the source dataset $v_\mathcal{S}^i$ and a video from the target dataset $v_\mathcal{T}^j$, we define the similarity between $v_\mathcal{S}^i$ and $v_\mathcal{T}^j$ as:
\begin{equation}\label{eq:sm}
    K_{ji} = \Phi(v_\mathcal{S}^i, v_\mathcal{T}^j),
\end{equation}
where the output of the similarity function $\Phi$ is a scalar.

\begin{figure}[t!]
\centering
   \includegraphics[width=1.\linewidth]{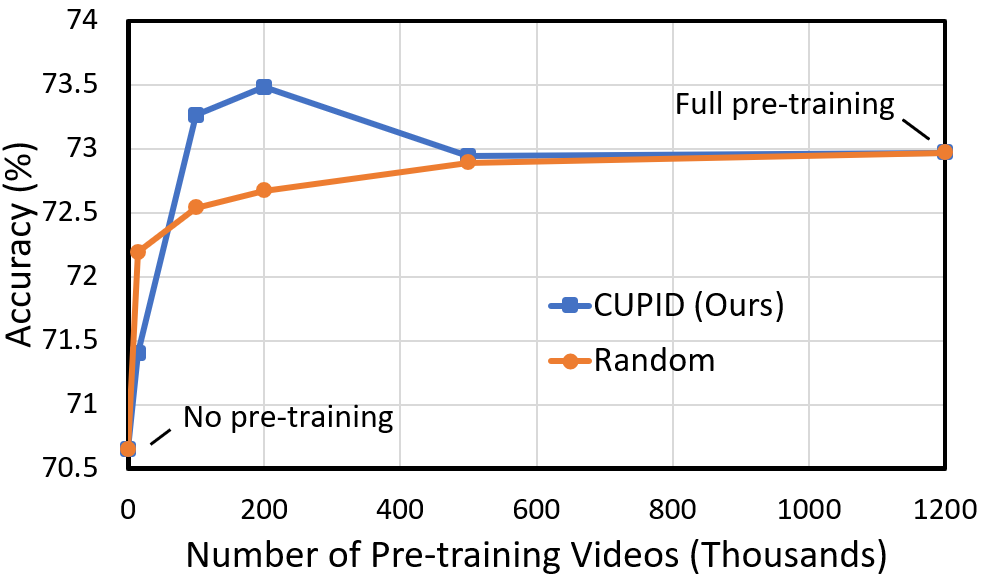}
   \caption{Results on TVQA~\cite{lei2018tvqa} validation set w.r.t. the size of pre-training data sampled from HowTo100M~\cite{miech2019howto100m}. The proposed \textsc{Cupid} method ({\color{blue}\textbf{blue}}) uses task-adaptive sampling of pre-training data, while the baseline ({\color{orange}\textbf{orange}}) uses random sampling. With random sampling, we see a slow performance climbing followed by a saturation after showing the model 500K videos. Our proposed method peaks early at 200K videos and yields higher accuracy.}
\label{fig:tvqa_c}
\end{figure}

Since instructional videos usually consist of multiple scenes with varying appearances and activities, a single embedding vector for the entire video often fails to capture the rich semantics of the full video. We therefore decompose each video into clips, with the start and end timestamps determined either by subtitle boundaries, or uniform segmenting intervals, or manually-annotated event boundaries provided in the target datasets~\cite{ZhXuCoAAAI18}.
Let $Q$ be the number of clips in $v_\mathcal{S}^i$ and $L$ be the number of clips in $v_\mathcal{T}^j$.
We project all the clips into the same embedding space using a pre-trained feature extractor to obtain an embedding vector of dimension $d$ for each clip (note that the feature vector does not necessarily have to be normalized to have $l_2$ norm equal to one). Thus, the final representation for video $v_\mathcal{S}^i$ with clip embedding vectors stacked horizontally as a matrix $M_\mathcal{S}^i\in \mathbb{R}^{d\times Q}$. Similarly, $v_\mathcal{T}^j$ is represented as $M_\mathcal{T}^j\in \mathbb{R}^{d\times L}$). Eq.~\ref{eq:sm}
is now computed as:
\begin{equation}
    K_{ji} = \text{Mean}((M_\mathcal{T}^j)^\top M_\mathcal{S}^i),
\end{equation}
where $\text{Mean}(\cdot)$ performs an element-wise average on the input matrix. We also experiment with other trivial pooling strategies such as max pooling, which yields inferior empirical results.
A similarity matrix $K\in \mathbb{R}^{P\times N}$ can be constructed in this way, with $K_{ji}$ representing the similarity score of the video pair $(v_\mathcal{S}^i, v_\mathcal{T}^j)$.

\begin{table}[t]
\centering
\small
\begin{tabular}{llcccc}
\toprule
\multirow{2}{*}{Method} & \multirow{2}{*}{Curated Data} & \multicolumn{3}{c}{TVR} & TVQA \\ \cline{3-6} 
 &  & R@1 & R@10 & R@100 & Acc. \\
 \midrule
HERO & None & 2.98 & 10.65 & 18.25 & 70.65 \\
 & Random   & 5.20 & 17.01 & 24.54 & 74.35 \\ 
 \midrule
\textsc{Cupid} & KNN & \textbf{5.55} & 17.60 & 25.65 & 74.67 \\ 
 & Avg. Sim. & \textbf{5.55} & \textbf{17.61} & \textbf{25.73} & \textbf{74.92} \\ 
\bottomrule
\end{tabular}
\vspace{2mm}
\caption{Results on out-of-domain tasks with our proposed SAP.}\label{tab:sap_results}
\end{table}

\paragraph{Curation Strategies.} Two strategies are adopted to populate the curated video clips, given a pre-defined pre-training data capacity $c$ ($c$ varies from 1.2\% to the full dataset; we experiment on its impact on downstream performance later in this section).

$(i)$ \emph{Averaged Similarity (Avg. Sim.).} We mean-pool the similarity matrix $K$ by column to obtain the averaged similarity score between each source video and the entire target dataset. Videos with the top $c$ values are selected. This filtering strategy relies on an overall ``visual impression'' of the target data.

$(ii)$ \emph{K-Nearest Neighbours (KNN).} For each row in $K$, we select the top values according to a fixed number
to ensure the total number of selected source videos is $c$. This filtering strategy is instance-aware and requires more rigorous combing than curation by Avg. Sim. We observe in our experiments that this over-discrimination negatively impacts data diversity. As a trade-off, we loosen the constraint by first obtaining a pool 2-4x larger than the target $c$ 
, then randomly select $c$ videos from this pool. 

Note that each video clip is usually associated with either a subtitle or a human-annotated caption, so the same methodology is applicable to text embeddings or video-text embeddings. However, the accompanying text is not guaranteed (\eg, \cite{kuehne2011hmdb}) and is not discussed in this work.

\paragraph{More Advanced Curation Strategies.} Inspired by incremental learning, where training samples are ranked by their difficulty and are learned in an incremental fashion, we propose a similar approach for data curation. We first rank each pre-training video according to their similarity to the downstream dataset by using the two aforementioned strategies. Then, the model is exposed to top $c$ training samples in a incremental fashion, where $c$ is decreasing throughout the training (\eg, from 200k to 15k). Despite the sophistication of this approach, we do not observe any significant empirical boost compared to other simpler methods. Therefore, we leave the discussion for the Appendix.

\paragraph{Pre-processing and Pre-training.} Clip boundaries during video decomposition are defined as follows. For YouCook2, clip boundaries are determined by the ground-truth event segments from the corresponding training set. For HMDB51, which consist of much shorter videos, we uniformly divide a whole video into 20 windows. For TV and HowTo100M dataset, clips are determined by subtitle boundaries.
For a fair comparison to all baseline methods,
we preserve the same pre-training \& fine-tuning strategy, hyper-parameters, and optimizers, unless otherwise specified in Section~\ref{sec:sota}, to exclude factors other than the pre-training data itself. For all experiments in our work, the pre-training data is carefully filtered to exclude overlapping with any downstream data.

\begin{table}[t]
\centering
\small
\begin{tabular}{lcccccccc}
\toprule
 Method & B@3 & B@4 & M & C \\
 \midrule
 15K HT100M (Random) & 12.67 & 7.92 & 15.18 & 98.67 \\ 
 \midrule
 15K HT100M (Heuristic) & 13.93 & 8.83 & 16.10 & 105.94\\ 
 15K HT100M (KNN) & 14.19 & 9.13 & 16.14 & 106.24 \\ 
 15K HT100M (Avg. Sim.) & \textbf{14.42} & \textbf{9.34} & \textbf{16.47} & \textbf{110.45} \\ 
\bottomrule
\end{tabular}
\vspace{2mm}
\caption{Comparison between HAP and SAP on YouCook2 captioning.}\label{tab:hap_vs_sap_results}
\end{table}

\begin{table*}[t]
\centering
\small
\begin{tabular}{lllll}
\toprule
Downstream Task & Domain & Method & Feature Extractor & Our Curated Pre-training \\
\midrule
TVR and TVQA & Out-of-domain & HERO & ImageNet/Kinetics pre-trained & 200k from HowTo100M and TV \\
HMDB51 & Near-domain & \textsc{Cupid-clip} & WIT~\cite{radford2021learning} pre-trained backbone & 15k from HowTo100M \\
YouCook2 captioning & In-domain & \textsc{Cupid-vlp} & HowTo100M pre-trained & 15k from HowTo100M \\
YouCook2 retrieval & In-domain & MIL-NCE & HowTo100M pre-trained backbone & 15k from HowTo100M \\
\bottomrule
\end{tabular}
\vspace{2mm}
\caption{A summary of downstream tasks, domain genres, methods, and pre-training settings.}\label{tab:summary}
\end{table*}

\paragraph{Results on Out-of-domain Tasks.} 
We start with an analysis on the effect of pre-training data size $c$, with experiments on TVQA by varying the pre-training data size from 15K (1.2\%) to 1200K (full). Results are shown in Figure~\ref{fig:tvqa_c}.
Our proposed method (SAP with Avg. Sim. curation) is denoted in blue and the baseline with random sampling is in orange.
For a fair comparison, we run all the experiments under the same training configuration. We set the same number of training steps (100K, or equivalent to 70 epochs on full HowTo100M) for all experiments and ensure training convergence. Results show that our proposed method  only underperforms the baseline when data is extremely scarce (\ie, 15K), possibly due to low data diversity. As the data size $c$ increases, our method catches up quickly and outperforms the baseline. It achieves the best performance at $c=200\text{K}$, which is 4\% relatively better than the model without pre-training, even outperforming the model that has seen 5x more data. This advantage diminishes when we further increase $c$, as eventually the subset becomes the full dataset.

Following convention from prior work~\cite{li2020hero}, we further merge the TV dataset to train our best performing model (\ie, Avg. Sim. at $c=200\text{K}$) for a fair comparison. Results are shown in Table~\ref{tab:sap_results}. Both Avg. Sim. and KNN outperform the random sampling baseline. On TVQA, our methods lead the baseline slightly. On TVR, the gap is widened to 3.5-7\% relatively. Another observation is that Avg. Sim. has a slight edge over KNN, indicating that curation based on an ``overall impression'' of the target data tends to gain a good balance between data relevance and diversity, hence better at discovering useful data to close the domain gap.

\paragraph{Comparison Between HAP and SAP.} We conduct experiments on the in-domain task YouCook2 captioning to put HAP and SAP in a side-by-side comparison (Table~\ref{tab:hap_vs_sap_results}). To maintain a fair setting on pre-training data size $c$, we set $c=15\text{K}$ in all methods to be consistent with the data size obtained from heuristic curation. Key observations are: Avg. Sim. works consistently the best across tasks; our best performing method Avg. Sim. brings 18-30\% over random sampling baseline and 2-6\% over HAP. 

\subsection{Comparison with SotA}\label{sec:sota}

Among the numerous video-language pre-training methods proposed recently, we adopt two competitive open-sourced models: HERO~\cite{li2020hero}, which is reconstructive\footnote{HERO uses video-subtitle matching (VSM), so it is not strictly reconstructive. But considering that it is based on BERT, and MLM is its core objective, here we use it to represent ``reconstructive'' methods.} and MIL-NCE~\cite{miech2020end}, which is contrastive. 
On certain datasets (\eg, HMDB51), surprisingly, we observe that MIL-NCE severely underperforms its image counterpart CLIP pre-trained on image-text corpora~\cite{radford2021learning} (54.9 \vs 60.7 in accuracy), despite that the latter only takes in a single frame of each video. We therefore replace the S3D backbone from MIL-NCE with the Visual Transformer~\cite{dosovitskiy2020image} in CLIP. We further make modifications to CLIP such that it can take in multi-frame input. We name this method \head{\textsc{Cupid-clip}}.
\textsc{Cupid-clip} deploys the WebImageText (WIT)~\cite{radford2021learning} pre-trained ViT-B/32 as the backbone. To support multi-frame input, we sample frames from each video clip at a fixed frame rate (in our case, 16 frames at 5 fps, same as in MIL-NCE), patchify each frame into 32x32 patches, and then sequentially feed patches from each frame into the Transformer model. Note that the context input is now 16x longer compared to CLIP and the positional embedding needs to adjust accordingly. We initialize the positional embeddings for each frame separately from CLIP and finetune them during training such that the model can learn to leverage temporal positional information. During inference, we adopt the same strategy from MIL-NCE that aggregates clip embedding output from multiple windows to further boost the performance.

The aforementioned methods lean heavily towards understanding-based tasks, such as video QA and retrieval. To assess our proposed framework on generation-based tasks, there are few open-sourced solutions in the video domain. We resort to an alternative in the image domain, VLP~\cite{zhou2020unified}, one of the best-performing methods on image captioning. 
We name our method \head{\textsc{Cupid-vlp}}.
\textsc{Cupid-vlp} accepts segment-level features~\cite{miech2020end} extracted from uniformly divided video fragments, which empirically yields better results than frame-wise region proposal features in the original VLP. We sample 20 uniform windows in each video. For pre-training,  the misalignment issue between video clip and ASR subtitle is pronounced when the clip is short. To alleviate this issue, we concatenate every 3 consecutive subtitles as one training sample. 

A summary of downstream tasks, methods, and pre-training settings see Tab.~\ref{tab:summary}. 


\begin{table*}[t]
\centering
\small
\begin{tabular}{llcccccccc}
\toprule
\multirow{2}{*}{Method} & \multirow{2}{*}{Curated PT Data} & \multicolumn{4}{c}{Micro-Averaging} & \multicolumn{4}{c}{Macro-Averaging} \\ \cline{3-10} 
 & & B@3 & B@4 & M & C & B@3 & B@4 & M & C \\
 \midrule
Masked Trans.~\cite{zhou2018end} & None & 7.53 & 3.85 & 10.68$^\dagger$ & 37.9 & - & 1.42 & 11.20 & - \\
VideoBERT~\cite{sun2019videobert} & 312K videos & 7.59 & 4.33 & 11.94 & 55.0 & - & 1.79 & 10.80 & - \\
ActBERT~\cite{zhu2020actbert} & Full 1200K HT100M & 8.66 & 5.41 & 13.30 & 65.0 & - & - & - & - \\
 \midrule
\textsc{Cupid-vlp} (Ours) & 15K HT100M & \textbf{14.42} & \textbf{9.34} & \textbf{16.47} & \textbf{110.45} & \textbf{12.64} & \textbf{6.37} & \textbf{17.07} & \textbf{117.99} \\ 
\bottomrule
\end{tabular}
\vspace{2mm}
\caption{Comparison of our method \textsc{Cupid-vlp} and SotA methods on YouCook2 video captioning. $^\dagger$Erratum. Initially reported in~\cite{sun2019videobert} as 11.55 and rerun with the official \href{https://github.com/tylin/coco-caption}{coco-caption} evaluation protocol.}\label{tab:yc2}
\end{table*}


\begin{table*}[t]
\small
\centering
\def\w{20pt}  
\begin{tabular}{llccccccccc}
\toprule
\multirow{2}{*}{Method} & \multicolumn{3}{c}{TVR} & TVQA & \multicolumn{4}{c}{YouCook2 Text-to-Video Retrieval} & HMDB51 \\ \cline{2-10} 
 & R@1 & R@10 & R@100 & Acc. & R@1 & R@5 & R@10 & mR ($\downarrow$) & Acc. \\
 \midrule
SotA~\cite{li2020hero,kim2020dense,miech2020end,alayrac2020self}$^\ddagger$ & 6.21 & 19.34 & 36.66 & 74.09 & 13.90 & 36.34 & 48.87 & 11 & \textbf{67.10} \\ 
\midrule
MIL-NCE~\cite{miech2020end} (reproduced)$^\mathparagraph$ & - & - & - & - & 14.68 & 37.32 & 49.97 & 11 & 54.90 \\
\quad\quad + YouCook2 FT & - & - & - & - & 15.80 & 40.34 & 54.10 & 8 & - \\
\midrule
\textsc{Cupid} (ours) & \textbf{7.09} & \textbf{20.35} & \textbf{40.53} & \textbf{74.21} & \textbf{17.67} & \textbf{43.21} & \textbf{57.06} & \textbf{7} & 65.45 \\ 
\bottomrule
\end{tabular}
\vspace{2mm}
\caption{Comparison of our method \textsc{Cupid} and SotA methods on TVQA, TVR, YouCook2 retrieval, and HMDB51. Results on TVQA and TVR are on the test-public set. Action recognition on HMDB51 is conducted with linear evaluation on the (public) test set. ``FT'' indicates fine-tuning. $^\ddagger$We re-evaluated MIL-NCE based on the \href{https://github.com/antoine77340/MIL-NCE_HowTo100M/blob/master/csv/validation_youcook.csv}{public split}. The numbers reported from the paper are based on a \textit{private} smaller split and hence unfairly better. $^\mathparagraph$Our reproduced MIL-NCE results under the zero-shot setting. See text for details. 
}\label{tab:sota}
\end{table*}

\paragraph{Implementations Details.}
The batch size in \textsc{Cupid-vlp} is set to 300 and 256 for pre-training and fine-tuning. The learning rate is set to 2e-4 and 5e-5, respectively, after a light hyper-parameter search. We perform the pre-training for 30 epochs.
For \textsc{Cupid-clip}, the model is optimized with the same contrastive loss from MIL-NCE, with batch size 160 and learning rate 1e-6 for 300 epochs.
For HERO, the only change we made is adjusting the number of training steps. We set it to $300\text{K}$ steps only for pre-training experiments in Table~\ref{tab:sap_results} and~\ref{tab:sota}. We finetune TVR for 10k \vs 5k in the paper, as we observed the latter does not always guarantee convergence. Our pre-training only takes 6 days \vs the original three weeks under the same hardware configuration. For MIL-NCE, compared to the original setting, we can fit at most a batch of 600 into a 8x 32GB V100 machine, versus 8192 from the original paper (with a massive 64 x 128GB Cloud TPU v3 cluster). On the other hand, existing work~\cite{chen2020simple} observes a positive correlation between batch size and performance for contrastive loss. To mitigate the performance degradation due to a smaller batch size, we propose a strategy called $\text{N}^2$-NCE that expands the negative set with weak negatives by always using all the mismatched pairs from the batch (the number of negatives is quadratic to the batch size; see Appendix for details).

\paragraph{Evaluation Results.}
Results on YouCook2 video captioning are shown in Tab.~\ref{tab:yc2}, all methods are benchmarked on the val set, as the test set is not publicly available. We exclude work such as~\cite{hessel2019case, huang-etal-2020-multimodal} that take in subtitles during inference as subtitles might not always be available. 
Our proposed method leads the current best model ActBERT by a wide margin and sets new state of the art.
Since our method \textsc{Cupid-vlp} leverages a HowTo100M pre-trained feature extrator, it also showcases that general pre-training and our curated pre-training could be complementary. 

For TVQA and TVR, we select the best model (Avg. Sim.) from our model validation (Tab.~\ref{tab:sap_results}) to compare with SotA methods on the official test-public set. Results are shown in Table~\ref{tab:sota}.  On TVR, \cupidabbr even outperforms the SotA model with full HowTo100M pre-training in~\cite{li2020hero} by 5-14\% relatively, with significantly 80\% fewer training data and 60\% fewer training steps. On TVQA, \cupidabbr outperforms both the base model HERO (test accuracy: 73.61) and the latest SotA method~\cite{kim2020dense}. As of the date of submission, our method tops both \href{https://competitions.codalab.org/competitions/22780#results}{TVR} and \href{https://competitions.codalab.org/competitions/20415#results}{TVQA} leaderboard.

Next, we compares results on YouCook2 text-to-video retrieval, where the base model is MIL-NCE.
We first want to draw attention to our reproduced MIL-NCE, where we adopt an improved inference strategy that samples the middle point of each video window, rather than the start~\cite{miech2020end}. This along with finetuning on YouCook2 brings up our base model a notch. \cupidabbr further widens the gap between our method and the current SotA to 27\% on R@1, 19\% on R@5, 17\% on R@10, all relatively.
Note that we exclude results from concurrent unpublished work such as~\cite{ging2020coot} because their val set only has 3.2k clips \vs ours from the \href{https://github.com/antoine77340/MIL-NCE_HowTo100M/blob/master/csv/validation_youcook.csv}{public split} with 3.4k clips. We are under an unfair disadvantage while still performing better (17.67 \vs 16.70 at R@1). 
Finally on HMDB51, our proposed method \textsc{Cupid-clip} is on par with the SotA method~\cite{alayrac2020self} which leverages more pre-training data and the extra audio modality. It demonstrates the effectiveness of curated video-language pre-training in improving standalone video representation, echoing the conclusion from general pre-training~\cite{miech2020end}.

\section{Conclusion}
We propose a new adaptive pre-training approach with domain-focused data curation for video-and-language representation learning. From  a probing study on pre-training transferability across video domains, we observe a clear domain gap between the source pre-training data and out-of-domain downstream data, which hinders source-to-target knowledge transfer and leads to inferior downstream performance. 
We propose an effective framework \cupidabbr to bridge this domain gap by filtering and adapting source data to the target domain, which achieves new state of the art on various tasks across multiple domains. 


{\small
\bibliographystyle{ieee_fullname}
\bibliography{data_curation}

\begin{thebibliography}{10}\itemsep=-1pt

\bibitem{alayrac2016unsupervised}
Jean-Baptiste Alayrac, Piotr Bojanowski, Nishant Agrawal, Josef Sivic, Ivan
  Laptev, and Simon Lacoste-Julien.
\newblock Unsupervised learning from narrated instruction videos.
\newblock In {\em Proceedings of the IEEE Conference on Computer Vision and
  Pattern Recognition}, pages 4575--4583, 2016.

\bibitem{alayrac2020self}
Jean-Baptiste Alayrac, Adri{\`a} Recasens, Rosalia Schneider, Relja
  Arandjelovi{\'c}, Jason Ramapuram, Jeffrey De~Fauw, Lucas Smaira, Sander
  Dieleman, and Andrew Zisserman.
\newblock Self-supervised multimodal versatile networks.
\newblock {\em arXiv preprint arXiv:2006.16228}, 2020.

\bibitem{alwassel2019self}
Humam Alwassel, Dhruv Mahajan, Lorenzo Torresani, Bernard Ghanem, and Du Tran.
\newblock Self-supervised learning by cross-modal audio-video clustering.
\newblock {\em arXiv preprint arXiv:1911.12667}, 2019.

\bibitem{anne2017localizing}
Lisa Anne~Hendricks, Oliver Wang, Eli Shechtman, Josef Sivic, Trevor Darrell,
  and Bryan Russell.
\newblock Localizing moments in video with natural language.
\newblock In {\em Proceedings of the IEEE international conference on computer
  vision}, pages 5803--5812, 2017.

\bibitem{brown2020language}
Tom~B Brown, Benjamin Mann, Nick Ryder, Melanie Subbiah, Jared Kaplan, Prafulla
  Dhariwal, Arvind Neelakantan, Pranav Shyam, Girish Sastry, Amanda Askell,
  et~al.
\newblock Language models are few-shot learners.
\newblock {\em arXiv preprint arXiv:2005.14165}, 2020.

\bibitem{caba2015activitynet}
Fabian Caba~Heilbron, Victor Escorcia, Bernard Ghanem, and Juan Carlos~Niebles.
\newblock Activitynet: A large-scale video benchmark for human activity
  understanding.
\newblock In {\em CVPR}, pages 961--970, 2015.

\bibitem{carreira2017quo}
Joao Carreira and Andrew Zisserman.
\newblock Quo vadis, action recognition? a new model and the kinetics dataset.
\newblock In {\em CVPR}, pages 6299--6308, 2017.

\bibitem{chen2011collecting}
David Chen and William~B Dolan.
\newblock Collecting highly parallel data for paraphrase evaluation.
\newblock In {\em Proceedings of the 49th Annual Meeting of the Association for
  Computational Linguistics: Human Language Technologies}, pages 190--200,
  2011.

\bibitem{chen2020simple}
Ting Chen, Simon Kornblith, Mohammad Norouzi, and Geoffrey Hinton.
\newblock A simple framework for contrastive learning of visual
  representations.
\newblock {\em arXiv preprint arXiv:2002.05709}, 2020.

\bibitem{chen2020uniter}
Yen-Chun Chen, Linjie Li, Licheng Yu, Ahmed~El Kholy, Faisal Ahmed, Zhe Gan, Yu
  Cheng, and Jingjing Liu.
\newblock Uniter: Universal image-text representation learning.
\newblock In {\em ECCV}, 2020.

\bibitem{Cordts2016Cityscapes}
Marius Cordts, Mohamed Omran, Sebastian Ramos, Timo Rehfeld, Markus Enzweiler,
  Rodrigo Benenson, Uwe Franke, Stefan Roth, and Bernt Schiele.
\newblock The cityscapes dataset for semantic urban scene understanding.
\newblock In {\em CVPR}, 2016.

\bibitem{damen2018scaling}
Dima Damen, Hazel Doughty, Giovanni Maria~Farinella, Sanja Fidler, Antonino
  Furnari, Evangelos Kazakos, Davide Moltisanti, Jonathan Munro, Toby Perrett,
  Will Price, et~al.
\newblock Scaling egocentric vision: The epic-kitchens dataset.
\newblock In {\em ECCV}, pages 720--736, 2018.

\bibitem{devlin2018bert}
Jacob Devlin, Ming-Wei Chang, Kenton Lee, and Kristina Toutanova.
\newblock Bert: Pre-training of deep bidirectional transformers for language
  understanding.
\newblock {\em arXiv preprint arXiv:1810.04805}, 2018.

\bibitem{dosovitskiy2020image}
Alexey Dosovitskiy, Lucas Beyer, Alexander Kolesnikov, Dirk Weissenborn,
  Xiaohua Zhai, Thomas Unterthiner, Mostafa Dehghani, Matthias Minderer, Georg
  Heigold, Sylvain Gelly, et~al.
\newblock An image is worth 16x16 words: Transformers for image recognition at
  scale.
\newblock {\em arXiv preprint arXiv:2010.11929}, 2020.

\bibitem{escorcia2019temporal}
Victor Escorcia, Mattia Soldan, Josef Sivic, Bernard Ghanem, and Bryan Russell.
\newblock Temporal localization of moments in video collections with natural
  language.
\newblock {\em arXiv preprint arXiv:1907.12763}, 2019.

\bibitem{fernando2017self}
Basura Fernando, Hakan Bilen, Efstratios Gavves, and Stephen Gould.
\newblock Self-supervised video representation learning with odd-one-out
  networks.
\newblock In {\em CVPR}, pages 3636--3645, 2017.

\bibitem{gao2017tall}
Jiyang Gao, Chen Sun, Zhenheng Yang, and Ram Nevatia.
\newblock Tall: Temporal activity localization via language query.
\newblock In {\em Proceedings of the IEEE international conference on computer
  vision}, pages 5267--5275, 2017.

\bibitem{gao2014jhu}
Yixin Gao, S~Swaroop Vedula, Carol~E Reiley, Narges Ahmidi, Balakrishnan
  Varadarajan, Henry~C Lin, Lingling Tao, Luca Zappella, Benjam{\i}n B{\'e}jar,
  David~D Yuh, et~al.
\newblock Jhu-isi gesture and skill assessment working set (jigsaws): A
  surgical activity dataset for human motion modeling.
\newblock In {\em Miccai workshop: M2cai}, volume~3, page~3, 2014.

\bibitem{Geiger2012CVPR}
Andreas Geiger, Philip Lenz, and Raquel Urtasun.
\newblock Are we ready for autonomous driving? the kitti vision benchmark
  suite.
\newblock In {\em CVPR}, 2012.

\bibitem{ging2020coot}
Simon Ging, Mohammadreza Zolfaghari, Hamed Pirsiavash, and Thomas Brox.
\newblock Coot: Cooperative hierarchical transformer for video-text
  representation learning.
\newblock {\em arXiv preprint arXiv:2011.00597}, 2020.

\bibitem{goyal2017making}
Yash Goyal, Tejas Khot, Douglas Summers-Stay, Dhruv Batra, and Devi Parikh.
\newblock Making the v in vqa matter: Elevating the role of image understanding
  in visual question answering.
\newblock In {\em Proceedings of the IEEE Conference on Computer Vision and
  Pattern Recognition}, pages 6904--6913, 2017.

\bibitem{gu2018ava}
Chunhui Gu, Chen Sun, David~A Ross, Carl Vondrick, Caroline Pantofaru, Yeqing
  Li, Sudheendra Vijayanarasimhan, George Toderici, Susanna Ricco, Rahul
  Sukthankar, et~al.
\newblock Ava: A video dataset of spatio-temporally localized atomic visual
  actions.
\newblock In {\em CVPR}, pages 6047--6056, 2018.

\bibitem{gururangan2020don}
Suchin Gururangan, Ana Marasović, Swabha Swayamdipta, Kyle Lo, Iz Beltagy,
  Doug Downey, and Noah~A. Smith.
\newblock Don't stop pretraining: Adapt language models to domains and tasks.
\newblock In {\em ACL}, 2020.

\bibitem{han2020self}
Tengda Han, Weidi Xie, and Andrew Zisserman.
\newblock Self-supervised co-training for video representation learning.
\newblock {\em arXiv preprint arXiv:2010.09709}, 2020.

\bibitem{he2020momentum}
Kaiming He, Haoqi Fan, Yuxin Wu, Saining Xie, and Ross Girshick.
\newblock Momentum contrast for unsupervised visual representation learning.
\newblock In {\em CVPR}, pages 9729--9738, 2020.

\bibitem{hessel2019case}
Jack Hessel, Bo Pang, Zhenhai Zhu, and Radu Soricut.
\newblock A case study on combining asr and visual features for generating
  instructional video captions.
\newblock In {\em CoNLL}, 2019.

\bibitem{huang-etal-2020-multimodal}
Gabriel Huang, Bo Pang, Zhenhai Zhu, Clara Rivera, and Radu Soricut.
\newblock Multimodal pretraining for dense video captioning.
\newblock In {\em Proceedings of the 1st Conference of the Asia-Pacific Chapter
  of the Association for Computational Linguistics and the 10th International
  Joint Conference on Natural Language Processing}, pages 470--490, Suzhou,
  China, Dec. 2020. Association for Computational Linguistics.

\bibitem{jang2017tgif}
Yunseok Jang, Yale Song, Youngjae Yu, Youngjin Kim, and Gunhee Kim.
\newblock Tgif-qa: Toward spatio-temporal reasoning in visual question
  answering.
\newblock In {\em Proceedings of the IEEE Conference on Computer Vision and
  Pattern Recognition}, pages 2758--2766, 2017.

\bibitem{jia2021scaling}
Chao Jia, Yinfei Yang, Ye Xia, Yi-Ting Chen, Zarana Parekh, Hieu Pham, Quoc~V
  Le, Yunhsuan Sung, Zhen Li, and Tom Duerig.
\newblock Scaling up visual and vision-language representation learning with
  noisy text supervision.
\newblock {\em arXiv preprint arXiv:2102.05918}, 2021.

\bibitem{kim2020dense}
Hyounghun Kim, Zineng Tang, and Mohit Bansal.
\newblock Dense-caption matching and frame-selection gating for temporal
  localization in videoqa.
\newblock {\em arXiv preprint arXiv:2005.06409}, 2020.

\bibitem{kim-2014-convolutional}
Yoon Kim.
\newblock Convolutional neural networks for sentence classification.
\newblock In {\em EMNLP}, pages 1746--1751, Doha, Qatar, Oct. 2014. Association
  for Computational Linguistics.

\bibitem{krishna2017dense}
Ranjay Krishna, Kenji Hata, Frederic Ren, Li Fei-Fei, and Juan Carlos~Niebles.
\newblock Dense-captioning events in videos.
\newblock In {\em ICCV}, pages 706--715, 2017.

\bibitem{kuehne2011hmdb}
Hildegard Kuehne, Hueihan Jhuang, Est{\'\i}baliz Garrote, Tomaso Poggio, and
  Thomas Serre.
\newblock Hmdb: a large video database for human motion recognition.
\newblock In {\em ICCV}, pages 2556--2563. IEEE, 2011.

\bibitem{lee2017unsupervised}
Hsin-Ying Lee, Jia-Bin Huang, Maneesh Singh, and Ming-Hsuan Yang.
\newblock Unsupervised representation learning by sorting sequences.
\newblock In {\em ICCV}, pages 667--676, 2017.

\bibitem{lei2020mart}
Jie Lei, Liwei Wang, Yelong Shen, Dong Yu, Tamara~L Berg, and Mohit Bansal.
\newblock Mart: Memory-augmented recurrent transformer for coherent video
  paragraph captioning.
\newblock {\em arXiv preprint arXiv:2005.05402}, 2020.

\bibitem{lei2018tvqa}
Jie Lei, Licheng Yu, Mohit Bansal, and Tamara~L Berg.
\newblock Tvqa: Localized, compositional video question answering.
\newblock In {\em EMNLP}, 2018.

\bibitem{lei2020tvr}
Jie Lei, Licheng Yu, Tamara~L Berg, and Mohit Bansal.
\newblock Tvr: A large-scale dataset for video-subtitle moment retrieval.
\newblock In {\em ECCV}, 2020.

\bibitem{li2020hero}
Linjie Li, Yen-Chun Chen, Yu Cheng, Zhe Gan, Licheng Yu, and Jingjing Liu.
\newblock Hero: Hierarchical encoder for video+ language omni-representation
  pre-training.
\newblock In {\em EMNLP}, 2020.

\bibitem{li2016tgif}
Yuncheng Li, Yale Song, Liangliang Cao, Joel Tetreault, Larry Goldberg,
  Alejandro Jaimes, and Jiebo Luo.
\newblock Tgif: A new dataset and benchmark on animated gif description.
\newblock In {\em Proceedings of the IEEE Conference on Computer Vision and
  Pattern Recognition}, pages 4641--4650, 2016.

\bibitem{liu2019roberta}
Yinhan Liu, Myle Ott, Naman Goyal, Jingfei Du, Mandar Joshi, Danqi Chen, Omer
  Levy, Mike Lewis, Luke Zettlemoyer, and Veselin Stoyanov.
\newblock Roberta: A robustly optimized bert pretraining approach.
\newblock {\em arXiv preprint arXiv:1907.11692}, 2019.

\bibitem{lu2019vilbert}
Jiasen Lu, Dhruv Batra, Devi Parikh, and Stefan Lee.
\newblock Vilbert: Pretraining task-agnostic visiolinguistic representations
  for vision-and-language tasks.
\newblock In {\em NeurIPS}, pages 13--23, 2019.

\bibitem{luo2020univilm}
Huaishao Luo, Lei Ji, Botian Shi, Haoyang Huang, Nan Duan, Tianrui Li, Xilin
  Chen, and Ming Zhou.
\newblock Univilm: A unified video and language pre-training model for
  multimodal understanding and generation.
\newblock {\em arXiv preprint arXiv:2002.06353}, 2020.

\bibitem{miech2020end}
Antoine Miech, Jean-Baptiste Alayrac, Lucas Smaira, Ivan Laptev, Josef Sivic,
  and Andrew Zisserman.
\newblock End-to-end learning of visual representations from uncurated
  instructional videos.
\newblock In {\em CVPR}, pages 9879--9889, 2020.

\bibitem{miech2019howto100m}
Antoine Miech, Dimitri Zhukov, Jean-Baptiste Alayrac, Makarand Tapaswi, Ivan
  Laptev, and Josef Sivic.
\newblock Howto100m: Learning a text-video embedding by watching hundred
  million narrated video clips.
\newblock In {\em ICCV}, pages 2630--2640, 2019.

\bibitem{oord2018representation}
Aaron van~den Oord, Yazhe Li, and Oriol Vinyals.
\newblock Representation learning with contrastive predictive coding.
\newblock {\em arXiv preprint arXiv:1807.03748}, 2018.

\bibitem{qian2020spatiotemporal}
Rui Qian, Tianjian Meng, Boqing Gong, Ming-Hsuan Yang, Huisheng Wang, Serge
  Belongie, and Yin Cui.
\newblock Spatiotemporal contrastive video representation learning.
\newblock {\em arXiv preprint arXiv:2008.03800}, 2020.

\bibitem{radford2021learning}
Alec Radford, Jong~Wook Kim, Chris Hallacy, Aditya Ramesh, Gabriel Goh,
  Sandhini Agarwal, Girish Sastry, Amanda Askell, Pamela Mishkin, Jack Clark,
  et~al.
\newblock Learning transferable visual models from natural language
  supervision.
\newblock {\em arXiv preprint arXiv:2103.00020}, 2021.

\bibitem{ranzato2014video}
MarcAurelio Ranzato, Arthur Szlam, Joan Bruna, Michael Mathieu, Ronan
  Collobert, and Sumit Chopra.
\newblock Video (language) modeling: a baseline for generative models of
  natural videos.
\newblock {\em arXiv preprint arXiv:1412.6604}, 2014.

\bibitem{rohrbach2015dataset}
Anna Rohrbach, Marcus Rohrbach, Niket Tandon, and Bernt Schiele.
\newblock A dataset for movie description.
\newblock In {\em Proceedings of the IEEE conference on computer vision and
  pattern recognition}, pages 3202--3212, 2015.

\bibitem{lsmdc}
Anna Rohrbach, Atousa Torabi, Marcus Rohrbach, Niket Tandon, Chris Pal, Hugo
  Larochelle, Aaron Courville, and Bernt Schiele.
\newblock Movie description.
\newblock {\em IJCV}, 2017.

\bibitem{rohrbach2017movie}
Anna Rohrbach, Atousa Torabi, Marcus Rohrbach, Niket Tandon, Christopher Pal,
  Hugo Larochelle, Aaron Courville, and Bernt Schiele.
\newblock Movie description.
\newblock {\em International Journal of Computer Vision}, 123(1):94--120, 2017.

\bibitem{rohrbach2012database}
Marcus Rohrbach, Sikandar Amin, Mykhaylo Andriluka, and Bernt Schiele.
\newblock A database for fine grained activity detection of cooking activities.
\newblock In {\em CVPR}, pages 1194--1201. IEEE, 2012.

\bibitem{rouditchenko2020avlnet}
Andrew Rouditchenko, Angie Boggust, David Harwath, Dhiraj Joshi, Samuel Thomas,
  Kartik Audhkhasi, Rogerio Feris, Brian Kingsbury, Michael Picheny, Antonio
  Torralba, et~al.
\newblock Avlnet: Learning audio-visual language representations from
  instructional videos.
\newblock {\em arXiv preprint arXiv:2006.09199}, 2020.

\bibitem{sanabria2018how2}
Ramon Sanabria, Ozan Caglayan, Shruti Palaskar, Desmond Elliott, Lo{\"\i}c
  Barrault, Lucia Specia, and Florian Metze.
\newblock How2: a large-scale dataset for multimodal language understanding.
\newblock {\em arXiv preprint arXiv:1811.00347}, 2018.

\bibitem{sharma2018conceptual}
Piyush Sharma, Nan Ding, Sebastian Goodman, and Radu Soricut.
\newblock Conceptual captions: A cleaned, hypernymed, image alt-text dataset
  for automatic image captioning.
\newblock In {\em Proceedings of the 56th Annual Meeting of the Association for
  Computational Linguistics (Volume 1: Long Papers)}, pages 2556--2565, 2018.

\bibitem{shin-etal-2020-biomegatron}
Hoo-Chang Shin, Yang Zhang, Evelina Bakhturina, Raul Puri, Mostofa Patwary,
  Mohammad Shoeybi, and Raghav Mani.
\newblock {B}io{M}egatron: Larger biomedical domain language model.
\newblock In {\em EMNLP}, pages 4700--4706, Online, Nov. 2020. Association for
  Computational Linguistics.

\bibitem{singh2020we}
Amanpreet Singh, Vedanuj Goswami, and Devi Parikh.
\newblock Are we pretraining it right? digging deeper into visio-linguistic
  pretraining.
\newblock {\em arXiv preprint arXiv:2004.08744}, 2020.

\bibitem{soomro2012ucf101}
Khurram Soomro, Amir~Roshan Zamir, and Mubarak Shah.
\newblock Ucf101: A dataset of 101 human actions classes from videos in the
  wild.
\newblock {\em arXiv preprint arXiv:1212.0402}, 2012.

\bibitem{srivastava2015unsupervised}
Nitish Srivastava, Elman Mansimov, and Ruslan Salakhudinov.
\newblock Unsupervised learning of video representations using lstms.
\newblock In {\em ICML}, pages 843--852, 2015.

\bibitem{sun2019learning}
Chen Sun, Fabien Baradel, Kevin Murphy, and Cordelia Schmid.
\newblock Learning video representations using contrastive bidirectional
  transformer.
\newblock {\em arXiv preprint arXiv:1906.05743}, 2019.

\bibitem{sun2019videobert}
Chen Sun, Austin Myers, Carl Vondrick, Kevin Murphy, and Cordelia Schmid.
\newblock Videobert: A joint model for video and language representation
  learning.
\newblock In {\em ICCV}, pages 7464--7473, 2019.

\bibitem{tang2019coin}
Yansong Tang, Dajun Ding, Yongming Rao, Yu Zheng, Danyang Zhang, Lili Zhao,
  Jiwen Lu, and Jie Zhou.
\newblock Coin: A large-scale dataset for comprehensive instructional video
  analysis.
\newblock In {\em CVPR}, pages 1207--1216, 2019.

\bibitem{tapaswi2016movieqa}
Makarand Tapaswi, Yukun Zhu, Rainer Stiefelhagen, Antonio Torralba, Raquel
  Urtasun, and Sanja Fidler.
\newblock Movieqa: Understanding stories in movies through question-answering.
\newblock In {\em CVPR}, pages 4631--4640, 2016.

\bibitem{torabi2015using}
Atousa Torabi, Christopher Pal, Hugo Larochelle, and Aaron Courville.
\newblock Using descriptive video services to create a large data source for
  video annotation research.
\newblock {\em arXiv preprint arXiv:1503.01070}, 2015.

\bibitem{twinanda2016endonet}
Andru~P Twinanda, Sherif Shehata, Didier Mutter, Jacques Marescaux, Michel
  De~Mathelin, and Nicolas Padoy.
\newblock Endonet: a deep architecture for recognition tasks on laparoscopic
  videos.
\newblock {\em IEEE transactions on medical imaging}, 36(1):86--97, 2016.

\bibitem{xu2017video}
Dejing Xu, Zhou Zhao, Jun Xiao, Fei Wu, Hanwang Zhang, Xiangnan He, and Yueting
  Zhuang.
\newblock Video question answering via gradually refined attention over
  appearance and motion.
\newblock In {\em Proceedings of the 25th ACM international conference on
  Multimedia}, pages 1645--1653, 2017.

\bibitem{xu2016msr}
Jun Xu, Tao Mei, Ting Yao, and Yong Rui.
\newblock Msr-vtt: A large video description dataset for bridging video and
  language.
\newblock In {\em Proceedings of the IEEE conference on computer vision and
  pattern recognition}, pages 5288--5296, 2016.

\bibitem{yu2016video}
Haonan Yu, Jiang Wang, Zhiheng Huang, Yi Yang, and Wei Xu.
\newblock Video paragraph captioning using hierarchical recurrent neural
  networks.
\newblock In {\em Proceedings of the IEEE conference on computer vision and
  pattern recognition}, pages 4584--4593, 2016.

\bibitem{yu2018joint}
Youngjae Yu, Jongseok Kim, and Gunhee Kim.
\newblock A joint sequence fusion model for video question answering and
  retrieval.
\newblock In {\em Proceedings of the European Conference on Computer Vision
  (ECCV)}, pages 471--487, 2018.

\bibitem{yu2019activitynet}
Zhou Yu, Dejing Xu, Jun Yu, Ting Yu, Zhou Zhao, Yueting Zhuang, and Dacheng
  Tao.
\newblock Activitynet-qa: A dataset for understanding complex web videos via
  question answering.
\newblock In {\em Proceedings of the AAAI Conference on Artificial
  Intelligence}, volume~33, pages 9127--9134, 2019.

\bibitem{zhou2020unified}
Luowei Zhou, Hamid Palangi, Lei Zhang, Houdong Hu, Jason~J Corso, and Jianfeng
  Gao.
\newblock Unified vision-language pre-training for image captioning and vqa.
\newblock In {\em AAAI}, pages 13041--13049, 2020.

\bibitem{ZhXuCoAAAI18}
Luowei Zhou, Chenliang Xu, and Jason~J Corso.
\newblock Towards automatic learning of procedures from web instructional
  videos.
\newblock In {\em AAAI}, pages 7590--7598, 2018.

\bibitem{zhou2018end}
Luowei Zhou, Yingbo Zhou, Jason~J Corso, Richard Socher, and Caiming Xiong.
\newblock End-to-end dense video captioning with masked transformer.
\newblock In {\em CVPR}, pages 8739--8748, 2018.

\bibitem{zhu2020actbert}
Linchao Zhu and Yi Yang.
\newblock Actbert: Learning global-local video-text representations.
\newblock In {\em CVPR}, pages 8746--8755, 2020.

\bibitem{zhukov2019cross}
Dimitri Zhukov, Jean-Baptiste Alayrac, Ramazan~Gokberk Cinbis, David Fouhey,
  Ivan Laptev, and Josef Sivic.
\newblock Cross-task weakly supervised learning from instructional videos.
\newblock In {\em CVPR}, pages 3537--3545, 2019.

\end{thebibliography}
}

\newpage
\section{Appendix}
\subsection{Related Work}
Here, we further review the downstream tasks including text-to-video retrieval, video question answering, and video captioning in terms of task types, objectives, and datasets.

\head{Text-to-Video Retrieval.}
We categorize existing work on text-to-video retrieval into three types: Video Corpus Retrieval (VCR), Single Video Moment Retrieval (SVMR), and Video Corpus Moment Retrieval (VCMR). VCR is the most well-established task, where the goal is to retrieve a single video from a video corpus given a language query. Notable dataset examples serving this purpose include MSVD~\cite{chen2011collecting}, MSR-VTT~\cite{xu2016msr}, and YouCook2~\cite{ZhXuCoAAAI18}. SVMR is proposed more recently by Hendricks \etal~\cite{anne2017localizing} and Gao \etal~\cite{gao2017tall}. It retrieves a video moment (bounded by the start and end timestamps) from a given video that relates to the langusge query. DiDeMo~\cite{anne2017localizing}, ActivityNet Captions~\cite{krishna2017dense}, and Charades-STA~\cite{gao2017tall} are among the benchmarks for this task. Finally, VCMR came even more recently from \cite{escorcia2019temporal} and \cite{lei2020tvr} which extends the searching scope of SVMR to the entire video corpus.

\head{Video Question Answering.}
Overall, video question answering (VideoQA) addresses the video counterpart of the well-known VQA problem in the image domain~\cite{goyal2017making}. Given a video reference and a query question, the model needs to answer the question in terms of multiple choices~\cite{yu2018joint}, natural language (for open-ended questions)~\cite{xu2017video}, or fill-in-the-blank~\cite{rohrbach2017movie}. Depending on the duration of the video, VideoQA could target on short videos such as GIFs (\eg, TGIF-QA~\cite{jang2017tgif}), longer video clips (\eg, MSRVTT-QA~\cite{xu2017video} and MovieQA~\cite{tapaswi2016movieqa}), or an entire YouTube video (\eg, ActivityNet-QA~\cite{yu2019activitynet}). For long video contexts, sometimes a model needs to first ground the question to a related visual scene, before attempting the question (\eg, TVQA~\cite{lei2018tvqa}).

\head{Video Captioning.}
The aforementioned tasks are all understanding-based, aiming to understand the relationship between the visual modality and the language modality. For the sake of task diversity, we also consider a generation-based task: video captioning. The most standard video captioning task aims to describe a (short) video in a natural language sentence. Major benchmarks include MSVD~\cite{chen2011collecting}, MPII-MD~\cite{rohrbach2015dataset}, M-VAD~\cite{torabi2015using}, and MSR-VTT~\cite{xu2016msr}. When it comes to long videos, they often consist of multiple events with clear boundaries whose essence can hardly be captured by a single sentence. Therefore, there is a trend recently to first detect events out of a video and then render them into captions, or often called Dense Video Captioning. Representative datasets are ActivityNet Captions~\cite{krishna2017dense} and YouCook2~\cite{ZhXuCoAAAI18}. When event boundaries are provided beforehand, a task called video paragraph captioning is considered~\cite{yu2016video, lei2020mart}.

\subsection{Implementation Details}

\begin{figure}[!t]
\begin{center}
   \includegraphics[width=0.9\linewidth]{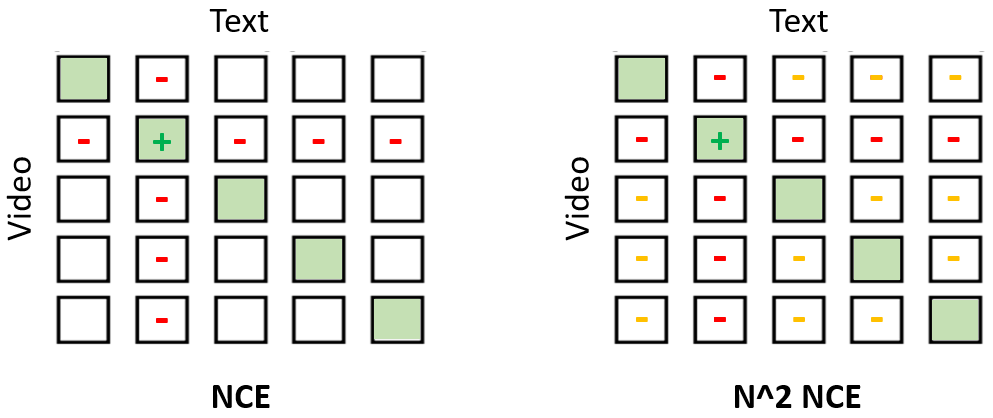}
\end{center}
   \caption{Comparison between NCE and our NCE with augmented weak negatives ($\text{N}^2$-NCE). In this example, the batch size is 5 and each cell represents a video-text pair. The diagonal indicates positive pairs and all others are negatives. For the video-text pair marked with {\color{green}green} ``{\color{green}+}'', in the original MIL-NCE, negatives are marked with {\color{red}red} ``{\color{red}-}''. In our $\text{N}^2$-NCE, we further leverage the weak negatives (marked by {\color{blue}blue} ``{\color{blue}-}'') to compensate our smaller batch size.}
\label{fig:n2_nce}
\end{figure}

\head{$\text{N}^2$-NCE.} In MIL-NCE and \textsc{Cupid-clip}, to mitigate the performance degradation due to a smaller batch size (mentioned in Sec. 4.3 of the main paper), we propose a strategy called $\text{N}^2$-NCE. $\text{N}^2$-NCE expands the negative set with weak negatives by always using all the mismatched pairs from the batch (see Fig.~\ref{fig:n2_nce}). Hence, with the augmented weak negatives, we expand the negative set from linear \emph{w.r.t.} the batch size to quadratic \emph{w.r.t.} the batch size. We set the learning rate to $1e-4$. Training takes roughly a day for 300 epochs on a 8x 32GB V100 machine. We further finetune the model on the training set of YouCook2 for 30 epochs.

\begin{table}[t]
\centering
\small
\begin{tabular}{lcccc}
\toprule
Curation Data & B@3 & B@4 & M & C \\
 \midrule
 15K HT100M (Random) & 11.05 & 5.03 & 15.82 & 106.28 \\ 
 \midrule
 15K HT100M (Heuristic) & 12.16 & 5.85 & 16.67 & 113.51 \\ 
 15K HT100M (KNN) & 12.22 & 5.80 & 16.73 & 114.77 \\ 
 15K HT100M (Avg. Sim.) & \textbf{12.64} & \textbf{6.37} & \textbf{17.07} & \textbf{117.99} \\ 
\bottomrule
\end{tabular}
\vspace{2mm}
\caption{Comparison between HAP and SAP on YouCook2 captioning.}\label{tab:hap_vs_sap_results2}
\end{table}

\subsection{More Results}

\paragraph{Comparison Between HAP and SAP.} Results on ``Macro-Averaging'' are in Table~\ref{tab:hap_vs_sap_results2}, in addition to results from Table~\ref{tab:hap_vs_sap_results}.


\paragraph{Data Curation with Incremental Learning.} Under this setup, we uniformly distribute the total number of training steps to curated dataset with descending sizes $c=\{200K, 100K, 15K\}$. On TVR, the R@1/10/100 are 5.44, 16.69, and 25.08 respectively. On TVQA, the accuracy is 74.50. On both tasks, this method underperforms the simpler static strategy Avg. Sim.

\end{document}